\documentclass{article}

    \PassOptionsToPackage{numbers, compress}{natbib}

\usepackage[preprint]{neurips_2023}

\usepackage[utf8]{inputenc} 
\usepackage[T1]{fontenc}    
\usepackage{hyperref}       
\usepackage{url}            
\usepackage{booktabs}       
\usepackage{amsfonts}       
\usepackage{nicefrac}       
\usepackage{microtype}      
\usepackage{xcolor}         

\usepackage{subcaption}
\usepackage{graphicx}
\usepackage{amsmath}
\usepackage{amssymb}
\usepackage{booktabs}
\usepackage{multirow}
\usepackage{algorithm}
\usepackage{algpseudocode}
\usepackage{color, colortbl}
\usepackage{subcaption}

\definecolor{Gray}{gray}{0.9}
\definecolor{Red}{RGB}{230, 57, 70}
\title{Boosting Human-Object Interaction Detection with Text-to-Image Diffusion Model}

\author{
  Jie Yang\textsuperscript{\rm 1,2}\thanks{co-first author}~~\thanks{This work was done when Jie Yang was intern at IDEA.}~~Bingliang Li\textsuperscript{\rm 1}$^{*}$~Fengyu Yang\textsuperscript{\rm 1}~Ailing Zeng\textsuperscript{\rm 2}\thanks{Corresponding author}~~ Lei Zhang\textsuperscript{\rm 2}~Ruimao Zhang \textsuperscript{\rm 1}$^\ddagger$ \quad \\ 
  \textsuperscript{\rm 1}Shenzhen Research Institute of Big Data, The Chinese University of Hong Kong, Shenzhen  \quad \\ 
  \textsuperscript{\rm 2}International Digital Economy Academy (IDEA)
}

\begin{document}

\maketitle

\begin{abstract}

This paper investigates the problem of the current HOI detection methods and introduces \texttt{DiffHOI}, a novel HOI detection scheme grounded on a pre-trained text-image diffusion model, which enhances the detector's performance via improved data diversity and HOI representation. 
We demonstrate that the internal representation space of a frozen text-to-image diffusion model is highly relevant to verb concepts and their corresponding context.
Accordingly, we propose an adapter-style tuning method to extract the various semantic associated representation from a frozen diffusion model and CLIP model to enhance the human and object representations from the pre-trained detector, further reducing the ambiguity in interaction prediction. 
%
Moreover, to fill in the gaps of HOI datasets, we propose \texttt{SynHOI}, a class-balance, large-scale, and high-diversity synthetic dataset containing over $140K$ HOI images with fully triplet annotations. It is built using an automatic and scalable pipeline designed to scale up the generation of diverse and high-precision HOI-annotated data. 
\texttt{SynHOI} could effectively relieve the long-tail issue in existing datasets and facilitate learning interaction representations. 
%
%
Extensive experiments demonstrate that \texttt{DiffHOI} significantly outperforms the state-of-the-art in regular detection (\textit{i.e.}, $41.50$ mAP) and zero-shot detection. Furthermore, \texttt{SynHOI} can improve the performance of
model-agnostic and backbone-agnostic HOI detection, particularly exhibiting
outstanding an $11.55\%$ mAP improvement in rare classes.

\end{abstract}

\section{Introduction}

Human-Object Interaction Detection (HOI), a core element in human-centric vision perception tasks such as human activity recognition~\cite{heilbron2015activitynet}, motion tracking~\cite{yi2022physical}, and anomaly behavior detection~\cite{liu2018future}, has attracted considerable attention over the past decades. 
HOI primarily involves localizing correlated human-object 2D positions within an image and identifying their interactions.
Despite numerous models proposed in recent years~\cite{hou2020visual,wan2019pose,liao2022gen,wu2022mining,zhang2021mining,zhou2022human}, practical implementation remains challenging due to the inherent ambiguity of verbs and subtle distinctions among interaction categories.

\begin{figure*}[t]
    \includegraphics[width=1\linewidth]{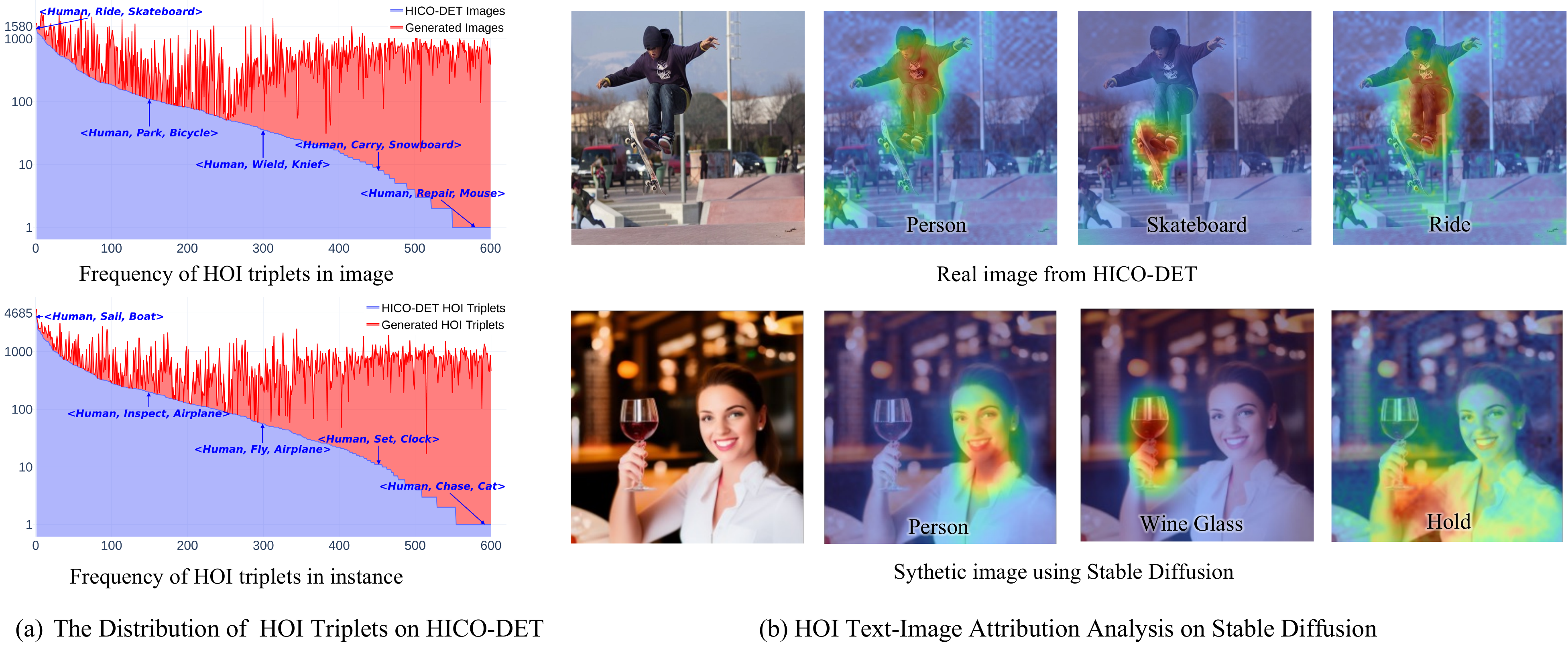}
        \vspace{-0.5cm}
    \caption{We show a) the long-tail distribution issue in HICO-DET and b) the high correlation between HOI text (\textit{i.e.}, nouns and verbs) and internal image features within the frozen stable diffusion. }
    \label{fig:intro}
    \vspace{-0.6cm}
\end{figure*}

We investigate the problems of the current methods from both data and model aspects.
From the data perspective, current datasets characterize an HOI instance as a ⟨\textit{human, object, interaction}⟩ triplet, and complete with comprehensive annotations, \textit{i.e.}, human and object bounding boxes and respective interaction types. 
We carefully analyze the widely-used HICO-DET dataset~\cite{chao:wacv2018}, which encompasses $38,118$ training images with $600$ distinct types of real-world HOI triplets, and have identified its several limitations that hinder the effective learning of interaction representations:
\textbf{(1) Class imbalance.} As demonstrated in Fig.~\ref{fig:intro}-(a), we observe it has a significant long-tail distribution in the HOI triplet types. This imbalanced label distribution is not conducive to learning, particularly when identifying interaction categories with only subtle variances. 
\textbf{(2) Small data size.} Upon further investigation, we discover that among its $600$ interaction categories, $51$ categories have only a single image available, and $435$ categories have less than $100$ images. Such limited data severely impacts the learning of these rare categories.
%
\textbf{(3) Limited diversity.} Compared with existing datasets, real-world scenarios typically exhibit more diversity due to the complex variation in human appearance attributes, environmental backgrounds, and shooting perspectives. This diversity causes a considerable decline in the performance of existing methods when the scene content changes.

From the model standpoint, the majority of existing HOI detectors, regardless of whether they employ two-stage~\cite{chao2018learning,gao2018ican,gupta2019no,hou2020visual,wan2019pose} or one-stage~\cite{liao2020ppdm,wang2020learning,kim2020uniondet,liao2022gen,wu2022mining,zhang2021mining,zhou2022human} strategies, are built upon pre-trained object detectors to enhance the initial localization of humans and objects. 
However, it remains challenging to carry out effective interaction prediction solely based on representations of humans and objects, sometimes with their spatial relationship. 
%
%
Specifically, verb concepts and their nuanced contextual information, including human posture, orientation, attention area and overall circumstances, can critically influence prediction.
Extracting this semantic information from data is typically inefficient and cumbersome, significantly limiting existing methods' performance.



Recently, text-to-image diffusion models~\cite{balaji2022ediffi,zhou2022towards,saharia2022photorealistic,ramesh2022hierarchical,rombach2022high} trained on massive internet-scale data have achieved significant performance in conditioned image generation, which provides high-quality, versatile, and semantically controllable image generation. 
Such an advantage offers us an effective means to generate rich and diverse realistic HOI images.
In practice, these models utilize cross-attention mechanisms between text embeddings and visual representations, signifying a substantial correlation between their feature spaces and semantic concepts in language. 
Inspired by this, we use DAAM~\cite{tang2022daam} to generate pixel-level linguistic concept heatmaps based on the image features of this diffusion model. 
As depicted in Fig.~\ref{fig:intro}-(b), beyond the noun concepts highlighted in prior studies~\cite{xu2023open,li2023guiding,shiparddiversity}, we find that the internal representation space of a frozen text-to-image diffusion model is highly relevant to verb concepts and their associated contexts.  
However, the challenge of extracting these verb-associated representations from the diffusion model for downstream HOI task still exist.

To address the aforementioned issues, we introduce \texttt{DiffHOI}, a novel HOI detection scheme based on text-to-image diffusion models (e.g., Stable Diffusion). For the first time, \texttt{DiffHOI} tries to leverage the generative and representative capabilities to benefit the HOI task, \textit{i.e.}, the extracted powerful verb-associated contextual representations of SD. \texttt{DiffHOI} consists of three components: the pre-trained human-object detector, interaction decoder, and object-interaction classifiers. We introduce an adapter-style tuning approach to align global and local semantic associated representations from the SD and CLIP model in the interaction decoder. These representations serve a critical role in comprehending the nuanced disparities in interactions and reducing interaction prediction ambiguity.

To fill in the shortcomings of existing long-tail HOI datasets, we present \texttt{SynHOI}, a class-balance, large-scale, and high-diversity synthetic HOI dataset with over $140K$ fully annotated HOI images, 
which can effectively facilitate learning interaction representations. To make the flow of dataset production scalable, we present an automatic pipeline, including the HOIPrompt design, automatic labeling and filtering, and quality verification, designed to continually scale up the generation of diverse and high-precision HOI-annotated data. 
Therefore, our contributions can be summarized as follows:
(1) We introduce a novel scheme, \texttt{DiffHOI}, which leverages both the generative and representation capacities of pre-trained text-to-image diffusion models to enhance the performance of HOI detection tasks.
%
(2) We present an automatic and scalable pipeline to generate realistic annotated HOI images characterized by varied attributes and scene contexts and propose a class-balance, large-scale, and high-diversity synthetic HOI dataset, namely \texttt{SynHOI}.
(3) Extensive experimental results demonstrate the proposed adapter-style tuning, together with the proposed SynHOI dataset, significantly improves the performance of HOI tasks under the regular and zero-shot settings and achieves the new state-of-the-art, \textit{i.e.}, $41.50$ mAP on HICO-DET.


\section{Related Work}
\textbf{HOI Detection.}
HOI detection task primarily encompasses three sub-problems, \textit{i.e.}, object detection, human-object pairing, and interaction recognition.
Previous HOI detectors can generally be divided into one-stage and two-stage paradigms.
The two-stage strategy employs an off-the-shelf detector to determine the locations and classes of objects, followed by specially-designed modules for human-object association and interaction recognition.
Most methods are dedicated to exploring additional feature streams to improve interaction classification, such as the appearance stream ~\cite{gao2018ican,li2019transferable,kim2020uniondet,hou2021detecting}, spatial stream~\cite{xu2019learning,bansal2020detecting,li2020detailed}, pose and body-parts stream~\cite{gupta2019no,wan2019pose,li2020detailed}, semantic stream~\cite{liu2020consnet,bansal2020detecting,gao2020drg}, and graph network~\cite{qi2018learning,xu2019learning,wang2020contextual,zhang2021spatially,park2023viplo}.
Instead, the one-stage strategy detects HOI triplets in a single forward pass by assigning human and object proposals to predefined anchors and then estimating potential interactions~\cite{liao2020ppdm, wang2020learning,kim2020uniondet,fang2021dirv}. 
Recently, the DETR-based HOI detectors~\cite{Tamura_2021_CVPR,tamura2021qpic,kim2021hotr,chen2021qahoi,ma2023fgahoi} have gained prominence in this paradigm, and they
formulate the HOI detection task as a set prediction problem, avoiding complex post-processing. In particular, many methods~\cite{zhang2021mining,liao2022gen,zhou2022human,chen2021reformulating} demonstrate promising performance improvements by disentangling
human-object detection and interaction classification as two decoders in a cascade manner. Our work builds on the top of the transformer-based HOI detection strategy and focuses on enhancing the learning of a dedicated interaction decoder.



\textbf{Zero-shot HOI Detection.}
Zero-shot HOI detection has emerged as a field aiming to identify unseen HOI triplet categories not present in the training data.
Previous research~\cite{bansal2020detecting,peyre2019detecting,peyre2019detecting,gupta2019no,hou2020visual,hou2021affordance,hou2021detecting,liu2020consnet} has addressed this task in a compositional manner, by disentangling the reasoning process on actions and objects during training. This approach enables the recognition of unseen HOI triplets during inference.
With the advancements in Vision-Language Models, such as CLIP~\cite{radford2021learning}, recent research~\cite{liao2022gen,wu2022end,ning2023hoiclip} has shifted focus toward transferring knowledge from CLIP to recognize unseen HOI concepts. This shift has resulted in a notable performance improvement in zero-shot settings. In this work, we aim to further explore the potential benefits of the Text-to-Image Diffusion model~\cite{rombach2022high} in enhancing zero-shot HOI detection.


\textbf{Diffusion Models for Image Generation and Vision Perception.}
Text-to-image diffusion models, such as DALLE-2~\cite{ramesh2022hierarchical}, Imagen~\cite{saharia2022photorealistic}, and Stable Diffusion~\cite{rombach2022high}, have shown considerable potentials in generating photorealistic images from free-form text prompts. This capability is attributed to the strong semantic correspondence between visual and language elements learned from a vast corpus of image-caption pairs. Recent research has proposed utilizing diffusion models to augment real datasets, assisting the training of downstream tasks~\cite{shiparddiversity,trabucco2023effective,bansal2023leaving,you2023diffusion}. Our work concentrates on generating synthetic HOI data to enhance HOI detection performance.
Moreover, text-to-image diffusion models pretrained on large-scale image-text pairs offer a high degree of control through customizable prompts. This aspect suits them for downstream tasks~\cite{xu2023open,li2023guiding}, where the additional feature representations from pretraining can bolster performance. 
This work first explores using a frozen text-image diffusion model for HOI detection.

\section{SynHOI-A Synthetic HOI Dataset}
\label{sec:data}
This section introduces how the proposed high-quality synthetic HOI dataset \texttt{SynHOI} is built via an automatic and scalable pipeline.

\subsection{Construction Process}
\begin{figure*}[t]
  \centering
  \begin{subfigure}[b]{1\textwidth}
  \centering
    \includegraphics[width=\textwidth]{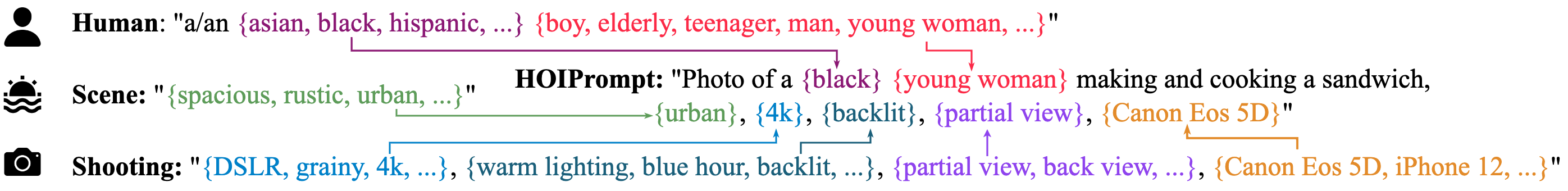}
    \caption{Examples of HOIPrompts}
    \label{fig:hoiprompt}
  \end{subfigure}
  \vfill
  \vspace{0.5cm}
  \begin{subfigure}[b]{1\textwidth}
    \includegraphics[width=\textwidth]{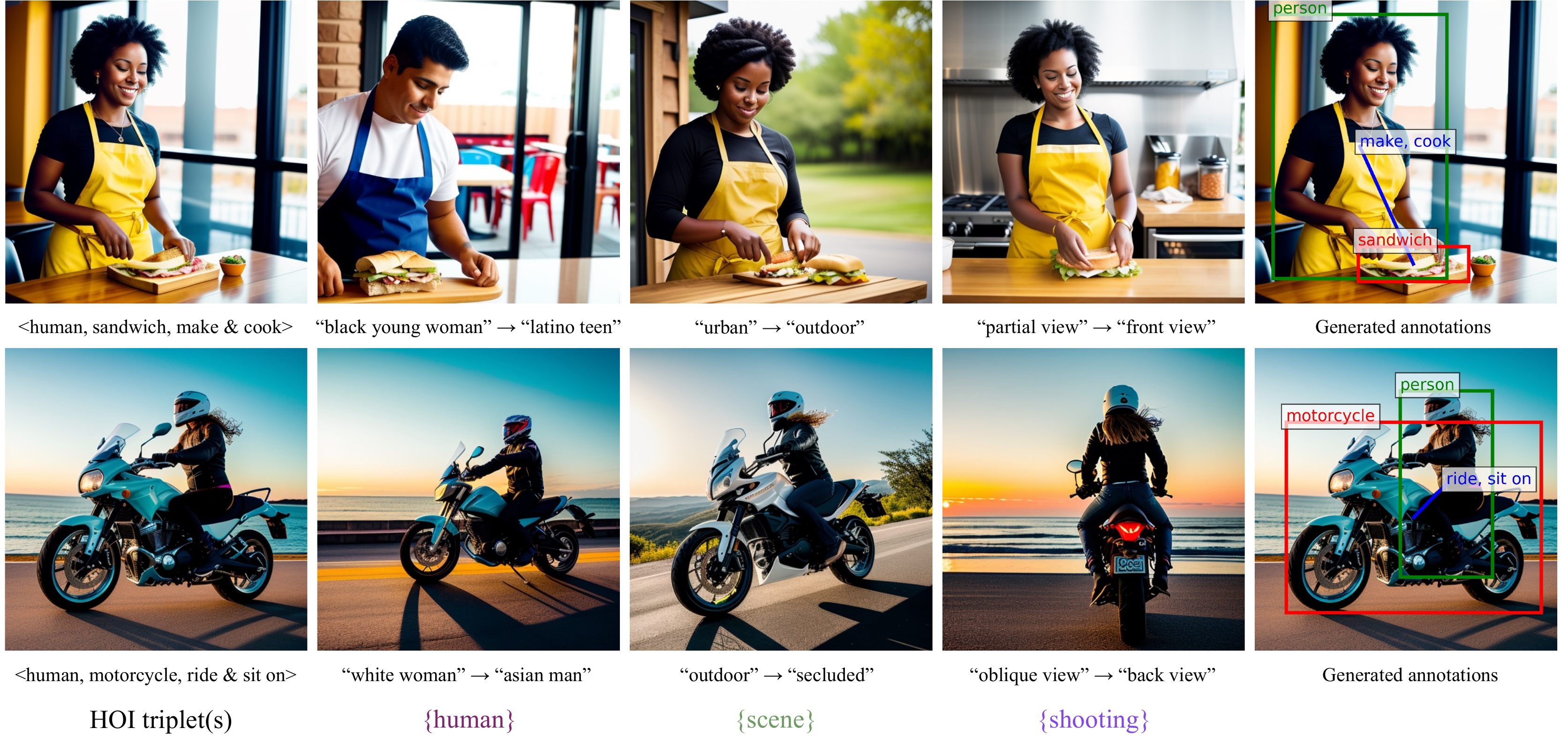}
    \caption{Examples of generated images and annotations}
    \label{fig:example_img}
  \end{subfigure}
    \vspace{-0.5cm}
  \caption{Illustration of a) HOIPrompts and b) how HOIPrompts guide the text-to-image generation process to enhance diversity. }
  \vspace{-0.8cm}
  \label{fig:data}
  
\end{figure*}

\textbf{The HOIPrompt design.} To address limitations in current datasets 
, we propose pre-defined HOIPrompts, as illustrated in Fig.~\ref{fig:data} (a).
Complete HOI triplets are formed by sampling verb and noun combinations from the HICO-DET dataset \cite{chao:wacv2018}, creating "a \{race\} \{age \& gender\} verb-ing an object."
To describe a person's appearance, we use the format "a \{race\} \{age \& gender\}," randomly selecting elements from the HOIPrompts. We generate images with combinations likely to occur together by analyzing co-occurring HOI triplets. For interaction environments, we select an adjective from a range of options to describe the atmosphere (denoted as "\{environment\}"). Photographic information is represented by four components ("\{quality\}," "\{lighting\}," "\{view\}," and "\{camera\}"), aligning the synthetic images with real HOI data and providing camera angle diversity. We further enhance diversity and quality through negative prompts and random model configurations. Overall, we generate $259,806$ images using HOIPrompts at this stage.

\textbf{Automatic Labeling and Filtering.} We design a three-step process to automatically annotate and filter the synthetic images. Firstly, we utilize a state-of-the-art detection model~\cite{zhang2022dino} to detect objects within each image. Secondly, we discard any images in which the confidence score of the detected object specified in the corresponding HOI triplet(s) prompt is below a threshold of $0.5$. Thirdly, we associate humans with objects in the images and assign the appropriate HOI category from the prompts to the human-object combination(s). In practice, we assign the HOI category to the person closest to the center of the object's bounding box. If not all humans in the image have an HOI label, we select the object closest to the corresponding human's bounding box. Upon completing the automated labeling and filtering process, we obtain a new synthetic dataset, namely \texttt{SynHOI}, including $146,772$ annotated images with complete HOI labels associated with the detected human-object interactions.



\textbf{Visualization and Manual Verification.} We develop a visualization tool to facilitate the manual inspection and filtering of any incorrect HOI annotations. We incorporate manual efforts to sample and inspect the annotated results to ensure their quality. During this inspection, we observe that the SOTA detector trained on COCO~\cite{lin2014microsoft} performs well in detecting humans and objects, indicating that the data distribution in \texttt{SynHOI} closely resembles that of natural images in COCO. However, due to the inherent ambiguity of verbs, a small number of synthetic interactions in \texttt{SynHOI} may be incompletely accurate, which also exists in the HICO-DET dataset.
To address this issue, we construct a subset comprising $5\%$ data of \texttt{SynHOI} (over $8K$), namely \texttt{SynHOI-Sub}, during the sampling and inspection process. This subset has undergone meticulous manual examination, resulting in annotations that are verified to be completely accurate.


\subsection{Data Characteristics}

\textbf{Large-scale.} \texttt{SynHOI} consists of $146,772$ images, $157,358$ person bounding boxes, $165,423$ object bounding boxes, and $282,140$ HOI triplet instances. It provides approximately four times the amount of training data compared to HICO-DET. 

\textbf{Class-balance.} \texttt{SynHOI} can effectively address the long-tail issue in previous datasets, where $343$ HOI categories have fewer than $50$ images in HICO-DET. Combining \texttt{SynHOI} with HICO-DET reduces the number of HOI categories with fewer than $50$ images to only three (refer to Fig.~\ref{fig:intro}-(a)).


\textbf{High-diversity.} 
\texttt{SynHOI} exhibits a high level of diversity, offering a wide range of visually distinct images. Fig.~\ref{fig:data}-(b) demonstrates the impact of random variations in people's descriptions, environments, and photographic information within the HOIPrompts on the diversity of synthetic images.

\textbf{High-quality.} \texttt{SynHOI} showcases high-quality HOI annotations. First, we employ CLIPScore~\cite{hessel2021clipscore} to measure the similarity between the synthetic images and the corresponding HOI triplet prompts. The \texttt{SynHOI} dataset achieves a high CLIPScore of $0.805$, indicating a faithful reflection of the HOI triplet information in the synthetic images. Second, Fig.~\ref{fig:data}-(b) provides evidence of the high quality of detection annotations in \texttt{SynHOI}, attributed to the effectiveness of the SOTA detector and the alignment of \texttt{SynHOI} with real-world data distributions. As mentioned earlier, we also release a carefully verified subset of \texttt{SynHOI}, called \texttt{SynHOI-Sub}.
\begin{figure*}[t]
    \centering
    \includegraphics[width=\linewidth]{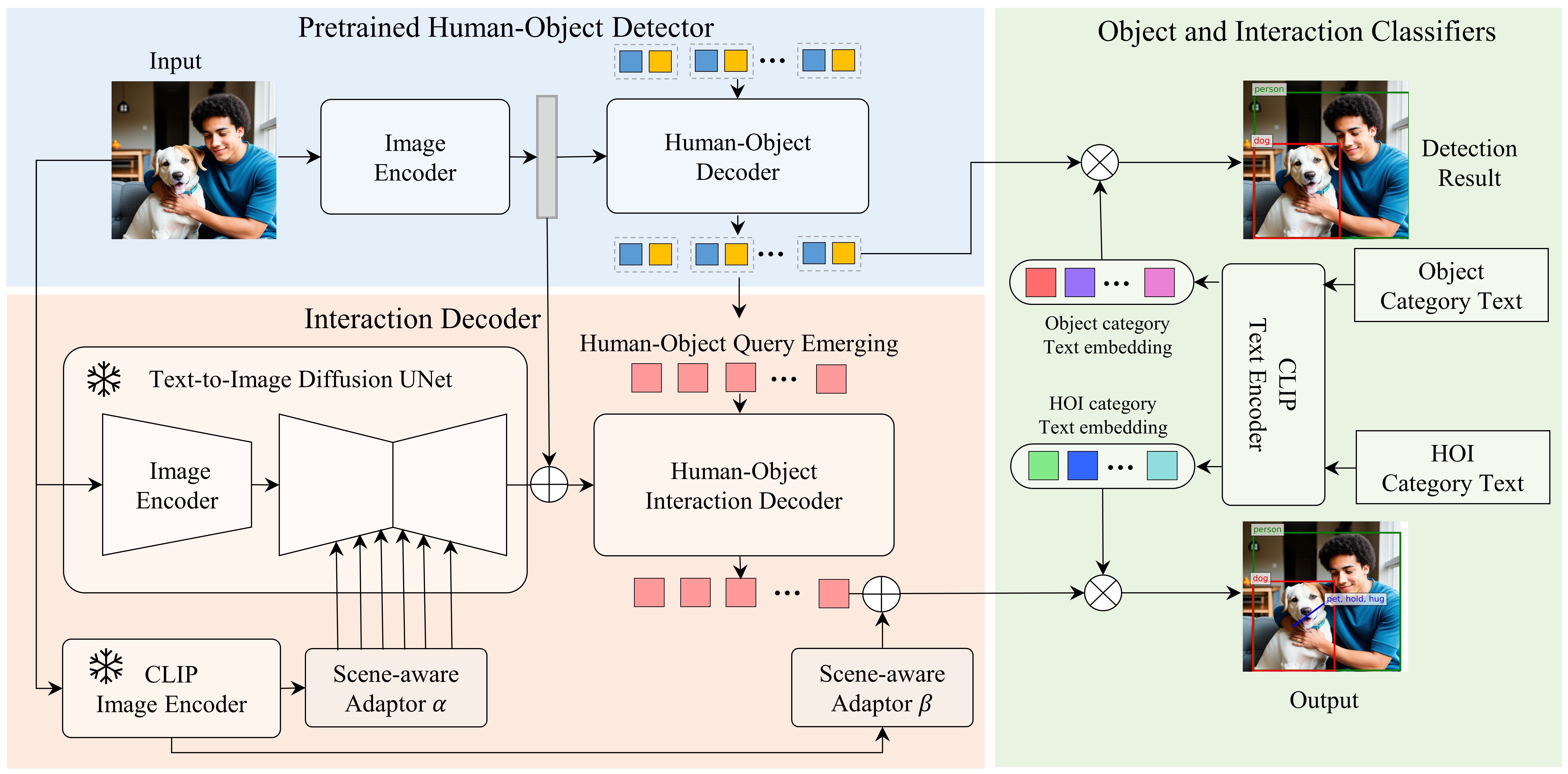}
    \caption{Overview of \texttt{DiffHOI}, comprising a pretrained human-object decoder, a novel interaction decoder, and CLIP-based object and interaction classifiers.  
    }
    \label{fig:framework}
    \vspace{-0.5cm}
\end{figure*}

\section{Methodology}
\subsection{Framework Overview}
\label{sec:overview}

The proposed \texttt{DiffHOI} framework is presented in Figure~\ref{fig:framework}, consisting of three components.
Primarily, we employ a transformer-based object detector $\textbf{D}_{obj}$~\cite{carion2020end} to extract local feature representations of humans and objects.
Using these extracted features, initial representations of interactions are computed and fed into the human-object interaction decoder, referred to as $\textbf{D}_{int}$, which subsequently updates these interaction representations.
Additionally, we employ the frozen Stable Diffusion (SD) Model $\textbf{F}_{sd}$\cite{rombach2022high} to generate semantically associated image feature maps, ensuring that distinct regions of these feature maps respond to various types of semantic information (see Sec.\ref{sec:StableDiffusion}).
%
%
Furthermore, we introduce the CLIP model~\cite{radford2021learning} and utilize its frozen image encoder $\textbf{F}_{clip}^{img}$ to extract comprehensive semantic representations that align with text descriptors. 
Two trainable scene-aware adaptors are introduced to adapt these semantic representations to suit the SD Model and interaction classifier (see Sec.~\ref{sec:CLIP}).
In practice, both the object classifier and interaction classifier are generated using the CLIP text encoder $\textbf{F}_{clip}^{text}$, which is applied during both the training and inference phases. 


%


\textbf{Pair-wise Human-Object Localization.} 
%
%
Given an input image $\boldsymbol{x}$, we first utilize the image encoder to extract feature maps $\boldsymbol{V}$, which is then adopted to localize humans and objects by using human-object decoder $\textbf{D}_{obj}$.
We further input pair-wise human-object queries $\boldsymbol{Q}_{\{h,o\}} \in \mathbb{R}^{2  \times N \times C }$ into the $\textbf{D}_{obj}$ and update them to $\boldsymbol{Q}^{\prime}_{\{h,o\}} \in \mathbb{R}^{2  \times N \times C }$, where $N$ is the number of paired queries and $C$ indicates the channel dimension. 
We use $\boldsymbol{Q}^{\prime}_{h,o}$ to predict the human bounding box $\boldsymbol{B}_{h} $ and object bounding box $\boldsymbol{B}_{o} $, whose dimension are ${ N \times 4 }$.
To predict the class label of each object, we further employ the text encoder $\textbf{F}_{clip}^{text}$ of CLIP to extract the object-aware embedding $\boldsymbol{T}^o_{clip} \in \mathbb{R}^{ K_1 \times C }$.
We further leverage $\boldsymbol{T}^o_{clip}$  to conduct the dot-product with $\boldsymbol{Q}^{\prime}_{o} \in \mathbb{R}^{N \times C }$ to predict the final object category distributions $\boldsymbol{P}_{o} \in \mathbb{R}^{ N \times K_1 }$, where $K_1$ denotes the total number of object classes.

\textbf{Human-Object Interaction Recognition.} 
We design a novel interaction decoder $\textbf{D}_{int}$ to perform better interaction understanding, where we exploit three types of visual representations as the input, \textit{i.e.}, the $\boldsymbol{V}$ from object detector, the $\boldsymbol{V}_{sd}$ from Stable Diffusion Model $\textbf{F}_{sd}$, and the $\boldsymbol{v}_{clip}$ from the image encoder $\textbf{F}_{clip}^{img}$ of CLIP. 
Specifically, we first perform average pooling on the human-object queries $\boldsymbol{Q}^{\prime}_{{h,o}}$, resulting in interaction queries $\boldsymbol{Q}{i} \in \mathbb{R}^{ N \times C }$.
Then, we feed $\boldsymbol{Q}_i$ into the interaction decoder $\textbf{D}_{int}$ to apply self-attention and subsequent cross-attention with the sum of $\boldsymbol{V}$ and $\boldsymbol{V}_{sd}$. This process updates the interaction queries from $\boldsymbol{Q}_{i}$ to $\boldsymbol{Q}^{\prime}_{i} \in \mathbb{R}^{ N \times C }$.
Moreover, we leverage $\boldsymbol{v}_{clip}\in \mathbb{R}^{C}$ to enhance each of the interaction query in $\boldsymbol{Q}^{\prime}_i$, performing element-wise addition.
To predict the interaction label of each human-object pair, we utilize the text encoder $\textbf{F}_{clip}^{text}$ of CLIP to extract the interaction-aware embedding $\boldsymbol{T}^i_{clip} \in \mathbb{R}^{ K_2 \times C }$.
Finally, we employ dot-product between $\boldsymbol{T}_{clip}^i$ and $\boldsymbol{Q}^{\prime}_i$ to predict the human-object interaction category distributions $\boldsymbol{P}_i \in \mathbb{R}^{ N \times K_2 }$, where $K_2$ is the number of interaction categories.


\subsection{Local Semantic Association via Stable Diffusion}
\label{sec:StableDiffusion}


%
%

The output features $\boldsymbol{V}$ from the object detector only retain human and object-oriented semantic information. 
In such a case, the interactional relationship and its corresponding context remain limited, resulting in sub-optimal interaction predictions.
%
Recenlty, Stable Diffusion~\cite{rombach2022high} (SD) has been primarily designed to produce high-quality images utilizing textual descriptors.
%
%
For the HOI task, we find that it be used to encourage diverse semantic information embedded in the textual representations to be associated with specific local regions of the images, as shown in Fig.~\ref{fig:intro}.
%
Inspired by the above observations, the feature maps derived from the SD model are expected to exhibit superior semantic correspondence within local regions, establishing an association between the semantic information of the text and the corresponding regions within the image. 
%
%
%
Accordingly, we utilize the UNet-based diffusion model to extract the feature maps, which are not only associated with the noun concepts, \textit{e.g.,} the human and objects, but also include the verb concepts and the corresponding contextual details.
%
%
Different from using a series of denoising steps to generate an image, we directly feed the input image $\boldsymbol{x}$ into the UNet (denoted as $\textbf{F}_{sd}$) and perform a single forward pass via the network. 
%
%
The output multi-scale image features are as $\boldsymbol{V}_{sd} = \textbf{F}_{sd}(\boldsymbol{x},\boldsymbol{A}_x)$, 
%
%
where $\boldsymbol{A}_x$ denotes the text-related representation that corresponds to $\boldsymbol{x}$.
Typically, $\boldsymbol{A}_x$ can be obtained by utilizing a text encoder, such as the CLIP text encoder, to encode the description of the image. 

However, as a discriminative task, the description of the HOI image is not available in the inference.
To address this, we replace it with the global feature representation of $\boldsymbol{x}$ extracted from the CLIP image encoder $\textbf{F}_{clip}^{img}$, which bears significant relevance to the textual representation. Please refer to Sec.~\ref{sec:CLIP} for more details.
%
%
%
%
As illustrated in the blocks in the red background of Fig.~\ref{fig:framework}, we proceed to combine the features $\boldsymbol{V}$ and $\boldsymbol{V}_{sd}$ to augment the representation ability of the verb concepts within $\boldsymbol{V}$ and use it to generate more informative $\boldsymbol{Q}^{\prime}_i$ for the final interaction prediction.

\subsection{CLIP Representation and Scene-aware Adaptor}
\label{sec:CLIP}

\textbf{CLIP Preliminaries.}
The CLIP model is trained to align visual and textual representations. It comprises an image encoder $\textbf{F}_{clip}^{img}$ and a text encoder $\textbf{F}_{clip}^{text}$, each followed by a linear layer, projecting image and text representations into a shared latent space.
Given an image $\boldsymbol{x}$, $\textbf{F}_{clip}^{img}$ extracts a global visual representation $\boldsymbol{v}_{clip} \in \mathbb{R}^{C}$, while $\textbf{F}_{clip}^{text}$ extracts a global text representation $\boldsymbol{T}_{clip} \in \mathbb{R}^{K \times C}$ for all $K$ predefined categories.
The prediction distribution over $K$ categories is then calculated as $\boldsymbol{S} =  \mathrm{softmax}(\boldsymbol{T}_{clip} * \boldsymbol{v}_{clip})$, where $*$ denotes the matrix-vector multiplication.

%
%


\textbf{Two Scene-aware Adaptors.} 
Given that $\boldsymbol{v}_{clip}$ represents a global feature, it inherently encompasses the contextual information of the overall scene.
As shown in Fig.~\ref{fig:framework}, we introduce two scene-aware adaptors, denoted as $\alpha$ and $\beta$, to project $\boldsymbol{v}_{clip}$ into feature spaces more consistent with the SD model and interaction predictor.
Regarding the scene-aware adaptor $\alpha$, since the CLIP model is trained to align global visual and textual representations within a latent space, the $\boldsymbol{v}_{clip}$ can be employed as a substitute for the textual representation. Hence, we can train an MLP to directly project $\boldsymbol{v}_{clip}$ to a textual space suitable for the SD model.
As for the scene-aware adaptor $\beta$, we project $\boldsymbol{v}_{clip}$ through it and incorporate the adapted version into each interaction query of $\boldsymbol{Q}^{\prime}_i$.
This adjustment allows for the tuning of these interaction queries to align more effectively with the CLIP-based classifiers $\boldsymbol{T}_{clip}^i$. 



\textbf{CLIP-based Classifiers.} 
Inspired by ~\cite{liao2022gen,wu2022end,ning2023hoiclip}, we use the CLIP text encoder $\textbf{F}_{clip}^{text}$ to generate both the object classifier and interaction classifier. 
Specifically, we initiate the process by transforming each object category or HOI category into a sentence using the hand-crafted template, 
\textit{i.e.} “A photo of a [Object]” or “A photo of a person [Verb-ing] a
[Object]”. 
Then these sentences can be encoded to obtain the object category embeddings $\boldsymbol{T}_{clip}^{o}$ or interaction category embeddings $\boldsymbol{T}_{clip}^{i} $.
Formally, the final object category distributions $\boldsymbol{P}_o$ and HOI category distributions $\boldsymbol{P}_i$ can be calculated as,
%
%
%
\begin{gather}
    \boldsymbol{P}_o = \mathrm{softmax}(\boldsymbol{Q}^{\prime}_{o} * {\boldsymbol{T}_{clip}^o}^{\intercal} ) \\
        \boldsymbol{P}_i = \mathrm{softmax}(\boldsymbol{Q}^{\prime}_{i} * {\boldsymbol{T}_{clip}^i}^{\intercal})
\end{gather}
where $\boldsymbol{Q}^{\prime}_{o} \in \mathbb{R}^{ N \times C }$ and $\boldsymbol{Q}^{\prime}_{o} \in \mathbb{R}^{ N \times C }$ denotes object queries and interaction queries, respectively, and $\mathrm{softmax}$ indicates the row-wise softmax operation.

\subsection{Loss Function}
\label{sec:train}
Following the query-based methods~\cite{liao2022gen,zhang2021mining}, we employ the Hungarian algorithm to match predictions to each ground-truth. The overall loss is computed between the matched predictions and their corresponding ground-truths, which includes the box regression loss $\mathcal{L}_{b}$, the intersection-over-union loss $\mathcal{L}_{g}$, the object class
loss $\mathcal{L}_{c}^{o}$
, and the interaction category loss $\mathcal{L}_{c}^{i}$,

\begin{equation}
    \mathcal{L} = \lambda_{b}\mathcal{L}_{b} +  \lambda_{g}\mathcal{L}_{g} + \lambda_{c}^{o} \mathcal{L}_{c}^{o} + \lambda_{c}^{i}\mathcal{L}_{c}^{i},
\end{equation}

where $\mathcal{L}_{b}$ and $\mathcal{L}_{g}$ contain both human and object localization. $\lambda_{b}$, $\lambda_{g}$, $\lambda_{c}^{o}$ and $\lambda_{c}^{i}$ are used to adjust the weights of each loss component. 

\section{Experiments}
\subsection{Experimental Settings}
Due to the page limit, we leave the detailed experimental settings in the Appendix.

\textbf{Datasets and Evaluation Metrics.} We evaluate our models on two widely used datasets: HICO-DET~\cite{chao2018learning} and V-COCO~\cite{gupta2015visual}. The mean Average Precision (mAP) is used as the evaluation metric, following standard protocols~\cite{zhang2021mining, liao2022gen}. 



\textbf{Zero-Shot Setting.} We conduct zero-shot experiments on HICO-Det, following the settings in~\cite{liao2022gen}: Rare First UC (RF-UC), Non-rare First UC (NF-UC), Unseen Object (UO) and Unseen Verb (UV).
\textbf{Implementation Details.}
We implement two variant architectures of \texttt{DiffHOI}: \texttt{DiffHOI}-S, and \texttt{DiffHOI}-L, where ‘S’ and
‘L’ refer to small and large, respectively. For \texttt{DiffHOI}-S, we use ResNet-50 as the backbone and a six-layer vanilla Transformer encoder~\cite{carion2020end} as the feature extractor. Both the human-object decoder and interaction decoder are three-layer vanilla Transformer decoders. For \texttt{DiffHOI}-L, we employ Swin-L~\cite{liu2021swin} as the backbone. In this variant, we replace all the transformer layers with deformable Transformer layers~\cite{zhang2022dino,zhu2020deformable}. 
We fine-tune the CLIP-based interaction and object classifiers
for regular settings with a small learning rate of $10^{-5}$ while freezing them for zero-shot settings. The diffusion model is frozen during training for all the settings.

\begin{table*}[t]
\begin{center}
\resizebox{1.\linewidth}{!}{
\begin{tabular}{llcccccccc}
\hline

\multirow{3}{*}{Method} & \multirow{3}{*}{Backbone} & \multicolumn{3}{c}{Default} && \multicolumn{3}{c}{Known Object} \\
\cmidrule{3-5} \cmidrule{7-9}
&& Full & Rare & Non-Rare && Full & Rare & Non-Rare   \\
\hline\hline
\multicolumn{5}{l}{\emph{Two-stage methods}} \\
STIP \cite{zhang2022exploring} & ResNet-50 & 32.22 & 28.15 & 33.43 && 35.29 & 31.43 & 36.45  \\
DEFR \cite{jin2022overlooked} & ViT-B/16 & 32.35 & 33.45 & 32.02 && - & - & -  \\
UPT \cite{zhang2022efficient} & ResNet-101 & 32.62 & 28.62 & 33.81 && 36.08 & 31.41 & 37.47  \\
ViPLO~\cite{park2023viplo} & ViT-B/32 & 34.95 & 33.83 & 35.28 && 38.15 & 36.77 & 38.56  \\
ViPLO~\cite{park2023viplo} & ViT-B/16 & 37.22 & 35.45 & 37.75 && 40.61 & 38.82 & 41.15 \\
\hline
\multicolumn{5}{l}{\emph{One-stage methods}} \\
QPIC \cite{tamura2021qpic}  & ResNet-101 & 29.90 & 23.92 & 31.69 && 32.38 & 26.06 & 34.27 \\
MSTR \cite{kim2022mstr} & ResNet-50 & 31.17 & 25.31 & 32.92 && 34.02 & 28.83 & 35.572 \\
SSRT \cite{iftekhar2022look}  & ResNet-101 & 31.34 & 24.31 & 33.32 && - & - & -  \\
CDN \cite{zhang2021mining}  & ResNet-101 & 32.07 & 27.19 & 33.53 && 34.79 & 29.48 & 36.38 \\
DOQ \cite{qu2022distillation} & ResNet-50 & 33.28 & 29.19 & 34.50 && - & - & -  \\
IF \cite{liu2022interactiveness} & ResNet-50 & 33.51 & 30.30 & 34.46 && 36.28 & 33.16 & 37.21\\
GEN-VLKT \cite{liao2022gen} & ResNet-50 & \underline{33.75}	& 29.25& 	35.10&&	\underline{36.78}	&32.75&	37.99 \\
QAHOI~\cite{chen2021qahoi} & Swin-T& 28.47 & 22.44 & 30.27 &&  30.99 & 24.83 &  32.84 \\
QAHOI~\cite{chen2021qahoi} & Swin-L& 35.78 & 29.80 & 37.56 && 37.59 & 31.66 & 39.36 \\
FGAHOI~\cite{ma2023fgahoi} & Swin-T& 29.94 &  22.24&  32.24 &&  32.48 & 24.16 & 34.97 \\
FGAHOI~\cite{ma2023fgahoi} & Swin-L& \underline{37.18} & 30.71 & 39.11  && \underline{38.93} & 31.93 & 41.02 \\
\rowcolor{Gray}\texttt{DiffHOI}-S                  & ResNet-50  & 34.41{\color{Red}$\uparrow_{1.96\%}$} & 31.07 & 35.40 &&  37.31{\color{Red}$\uparrow_{1.44\%}$} & 34.56 & 38.14 \\  
\rowcolor{Gray}\texttt{DiffHOI}-L                  & Swin-L &   40.63{\color{Red}$\uparrow_{9.28\%}$} & 38.10 & 41.38 && 43.14{\color{Red}$\uparrow_{10.81\%}$} & 40.24 &   44.01     \\     
\rowcolor{Gray}\texttt{DiffHOI}-L$^{\dagger}$                  & Swin-L & \textbf{41.50}{\color{Red}$\uparrow_{11.61\%}$} & \textbf{39.96} & \textbf{41.96} && \textbf{43.62}{\color{Red}$\uparrow_{12.05\%}$} & \textbf{41.41} & \textbf{44.28} \\    
\hline
\end{tabular}}
\end{center}

\caption{Performance comparison on HICO-DET in terms of mAP. $\dagger$ means pre-training using our \texttt{SynHOI} dataset. The \underline{underlined} highlights the compared results.} 
\label{tab:HICO-DET}
\vspace{-0.5cm}
\end{table*}
\subsection{Comparison to State-of-the-Art for Regular HOI Detection}
Tab.~\ref{tab:HICO-DET} and Tab.~\ref{tab:v-coco} present the performance comparison between \texttt{DiffHOI} and other state-of-the-art methods, which are grouped into
one-stage and two-stage methods. {For HICO-DET}, \texttt{DiffHOI}-L outperforms all existing one-stage and two-stage methods by a significant margin in all evaluation settings. Notably, it achieves a new state-of-the-art performance of $40.84$ mAP in the Default Full setting. Using the same backbone Swin-L, \texttt{DiffHOI}-L demonstrates a performance improvement of $3.45$ mAP compared to the current one-stage state-of-the-art method FGAHOI~\cite{ma2023fgahoi}. Furthermore, by incorporating our \texttt{SynHOI} dataset, the performance is further boosted to $41.50$ mAP.
For V-COCO, \texttt{DiffHOI}-L also surpasses the previous state-of-the-art methods with role AP scores of $65.7$ on Scenario 1 and $68.2$ on Scenario 2.

\begin{table*}[h]
    \begin{center}
                    \begin{minipage}{0.48\linewidth}
        \centering
        \resizebox{\linewidth}{!}{
            \makeatletter\def\@captype{table}\makeatother
\begin{tabular}{lcc}
\hline
Method    & AP (Scenario 1)  & AP (Scenario 2) \\ \hline\hline
\multicolumn{3}{l}{\emph{Two-stage methods}} \\
SCG \cite{zhang2021spatially}  &  54.2 & 60.9 \\
UPT \cite{zhang2022efficient}  &  61.3 & 67.1 \\
ViPLO~\cite{park2023viplo}& 60.9 & 66.6 \\
\hline
\multicolumn{3}{l}{\emph{One-stage methods}} \\ 
QPIC~\cite{tamura2021qpic} &    58.3 & 60.7 \\
MSTR \cite{kim2022mstr} & 62.0 & 65.2 \\
CDN \cite{zhang2021mining}  & 63.9 & 65.9 \\
IF \cite{liu2022interactiveness} &  63.0 & 65.2\\
GEN-VLKT~\cite{liao2022gen} &  62.4	& 64.5 \\
ParMap~\cite{wu2022mining}  &   63.0 & 65.1\\
\rowcolor{Gray}\texttt{DiffHOI}-S & 61.1 & 63.5   \\ 
\rowcolor{Gray}\texttt{DiffHOI}-L & \textbf{65.7} & \textbf{68.2}              
\\ \hline   
\end{tabular}
            }
\caption{{Performance comparison on V-COCO.}}
            \label{tab:v-coco}
        \end{minipage}
        \quad
        \begin{minipage}{0.48\linewidth}
        \vspace{-0.2cm}
        \centering
        \resizebox{\linewidth}{!}{
            \makeatletter\def\@captype{table}\makeatother
\begin{tabular}{lcccc}
\hline
Method                  & Type & Unseen         & Seen           & Full           \\
\hline\hline
GEN-VLKT\cite{liao2022gen}                & RF-UC          & 21.36          & 32.91          & 30.56          \\
\rowcolor{Gray}\texttt{DiffHOI}-S         & RF-UC    & 24.13 & 32.93 & 31.08     \\
\rowcolor{Gray}\texttt{DiffHOI}-L                  & RF-UC     & \textbf{28.76} & \textbf{38.01} & \textbf{36.16}     \\
\hline
GEN-VLKT\cite{liao2022gen}                & NF-UC          & 25.05          & 23.38          & 23.71          \\
\rowcolor{Gray}\texttt{DiffHOI}-S          & NF-UC   & 26.57 & 25.55 & 25.75      \\
\rowcolor{Gray}\texttt{DiffHOI}-L                  & NF-UC  & \textbf{29.45} &\textbf{31.68} &  \textbf{31.24} \\     
\hline
GEN-VLKT\cite{liao2022gen}                & UO             & \textbf{10.51}          & 28.92          & 25.63          \\
\rowcolor{Gray}\texttt{DiffHOI}-S          & UO    & 9.42 & 29.79 & 26.22    \\
\rowcolor{Gray}\texttt{DiffHOI}-L                  & UO  & 5.75 & \textbf{35.08} & \textbf{30.11}\\     
\hline
GEN-VLKT\cite{liao2022gen}                & UV             & 20.96          & 30.23          & 28.74          \\
\rowcolor{Gray}\texttt{DiffHOI}-S          & UV  & 23.10 & 30.91 & 29.72     \\
\rowcolor{Gray}\texttt{DiffHOI}-L                  &UV   &  \textbf{24.20}  & \textbf{36.81} & \textbf{35.04}\\      
\hline
\end{tabular}
            }
            \caption{Zero-shot performance comparison on HICO-DET. }
            \label{tab:zero-shot}
        \end{minipage}
    \end{center}
    \vspace{-0.5cm}
\end{table*}

\vspace{-0.3cm}
\subsection{Comparison to State-of-the-Art for Zero-Shot HOI Detection
}
\vspace{-0.2cm}
To comprehensively assess the zero-shot capability of \texttt{DiffHOI}, we conduct experiments on various zero-shot settings, as shown in Tab.~\ref{tab:zero-shot}. It demonstrates that \texttt{DiffHOI}-L outperforms the state-of-the-art method-GEN-VLKT~\cite{liao2022gen} in all zero-shot settings. Notably, \texttt{DiffHOI}-S with the same backbone achieves remarkable improvements, with a +$2.04$ mAP gain under the NF-UC setting for all categories and a +$2.14$ mAP improvement for rare categories under the UO setting.

\vspace{-0.4cm}
\subsection{Investigating the Usefulness of the SynHOI Dataset
}
\vspace{-0.2cm}
\begin{table*}[h]
    \begin{center}
        \begin{minipage}{0.48\linewidth}
        \centering
        \resizebox{\linewidth}{!}{
            \makeatletter\def\@captype{table}\makeatother
\begin{tabular}{lcccc}
\hline
Method                  & Backbone & Full         & Rare           & Non-Rare           \\
\hline\hline
GEN-VLKT$^{\ddagger}$~\cite{hou2020visual}        & ResNet50            &    33.04 & 29.10  &34.21       \\
\rowcolor{Gray}GEN-VLKT~\cite{hou2020visual}                    & ResNet50  &   34.43{\color{Red}$\uparrow_{4.21\%}$} & 32.46{\color{Red}$\uparrow_{11.55\%}$} & 35.02{\color{Red}$\uparrow_{2.37\%}$}      \\
\texttt{DiffHOI}-L                  & Swin-L          &  40.63 & 38.10 & 41.38      \\
\rowcolor{Gray}\texttt{DiffHOI}-L                & Swin-L          & 41.50{\color{Red}$\uparrow_{2.14\%}$} & 39.96{\color{Red}$\uparrow_{4.88\%6}$} & 41.96{\color{Red}$\uparrow_{1.40\%}$}        \\ 
\hline
\end{tabular}
            }
\caption{Effectiveness of pre-training using SynHOI and fine-tuning on HICO-DET. $^{\ddagger}$ indicates that our experiments reproduce the results.}
            \label{tab:pretraining}
        \end{minipage}
        \quad
                \begin{minipage}{0.48\linewidth}
        \centering
        \resizebox{\linewidth}{!}{
            \makeatletter\def\@captype{table}\makeatother
\begin{tabular}{lcccc}
\hline
Method                  & Backbone & Full         & Rare           & Non-Rare           \\
\hline\hline
GEN-VLKT$^{\ddagger}$~\cite{hou2020visual}        & ResNet50            &    33.04 & 29.10  &34.21       \\
\rowcolor{Gray}GEN-VLKT~\cite{hou2020visual}                    & ResNet50  & 33.73{\color{Red}$\uparrow_{2.09\%}$} &   30.08{\color{Red}$\uparrow_{3.37\%}$} & 34.82{\color{Red}$\uparrow_{1.78\%}$}              \\
\texttt{DiffHOI}-L                  & Swin-L          &   40.63 & 38.10 & 41.38       \\
\rowcolor{Gray}\texttt{DiffHOI}-L                & Swin-L          &     41.42{\color{Red}$\uparrow_{1.94\%}$} & 39.94{\color{Red}$\uparrow_{4.83\%}$} & 41.87{\color{Red}$\uparrow_{1.18\%}$}   \\ 
\hline
\end{tabular}
            }
\caption{Effectiveness of joint-training using SynHOI-sub and HICO-DET, evaluated on HICO-DET.}
            \label{tab:joint}
        \end{minipage}
    \end{center}
    \vspace{-0.7cm}
\end{table*}
As outlined in Sec.~\ref{sec:data}, we develop the \texttt{SynHOI} dataset, consisting of over $140K$ synthetic images that are generated, filtered, and annotated automatically. Additionally, we present a subset of \texttt{SynHOI}, named as \texttt{SynHOI-Sub}, which includes over $8K$ data that has undergone manual inspection to ensure the quality of annotations.
Here, we explore two strategies to leverage \texttt{SynHOI} and \texttt{SynHOI-Sub} to benefit the HOI detection task.

\textbf{Pre-training.} 
An intuitive strategy is to use \texttt{SynHOI} to pre-train an HOI detector and fine-tune the model on the target dataset. As shown in Tab.~\ref{tab:pretraining}, 
\texttt{SynHOI} could improve the performance of model-agnostic and backbone-agnostic HOI detection. 
Notably, there is a significant improvement in the rare AP when evaluating $138$ HOI categories with fewer than $10$ training instances in HICO-DET. Specifically, the rare AP increases significantly by $3.36$ mAP for GEN-VLKT and $1.86$ mAP for \texttt{DiffHOI}. It demonstrates that \texttt{SynHOI} can effectively relieve the long-tail issue and boost the interaction modeling in HICO-DET.

\textbf{Joint-training with HICO-DET.} \texttt{SynHOI-Sub} contains more diverse and class-balance images with accurate annotations than HICO-DET. We employ it in conjunction with HICO-DET to perform joint training of the models. As shown in Tab.~\ref{tab:joint}, the inclusion of \texttt{SynHOI-Sub} leads to an improvement of $0.69$ mAP for GEN-VLKT and $0.58$ mAP for \texttt{DiffHOI}, which can also contribute to improving the rare mAP and enhancing the learning of interactions within the rare categories.

\subsection{Ablation Study}
\vspace{-0.3cm}
\begin{table*}[h]
    \begin{center}
                            \begin{minipage}{0.4\linewidth}
    
        \resizebox{\linewidth}{!}{
            \makeatletter\def\@captype{table}\makeatother
\begin{tabular}{cc|ccc}
\hline
$\boldsymbol{V}_{sd}$ & $\boldsymbol{v}_{clip}$ &   Full & Rare & Non-rare                \\ \hline
& & 31.99 & 29.63 &  32.70                           \\
  \checkmark & &  32.92  & \textbf{31.29} & 33.41                                  \\
  \checkmark  &  \checkmark &     \textbf{34.41} & 31.07 & \textbf{35.40}                          \\
\hline
\end{tabular}
            }
\caption{{Ablation results of two key representations.}}
            \label{tab:representations}
        \end{minipage}
            \quad
                \begin{minipage}{0.4\linewidth}
        \resizebox{\linewidth}{!}{
            \makeatletter\def\@captype{table}\makeatother
\begin{tabular}{c|ccc}
\hline
Time step                   & Full         & Rare           & Non-Rare           \\
\hline
0                      & \textbf{34.41} & \textbf{31.07} & \textbf{35.40}      \\
100                   & 34.03 & 30.58 & 35.02              \\
500                    & 33.59 & 29.80 & 34.71            \\
\hline
\end{tabular}
            }
\caption{{Ablation results of different diffusion time steps.}}
            \label{tab:time-step}
        \end{minipage}
    \end{center}
    \vspace{-0.4cm}
\end{table*}

\textbf{Two Key Representations.} As described in Sec.~\ref{sec:overview}, we introduce $\boldsymbol{V}_{sd}$ derived from stable diffusion $\textbf{F}_{sd}$ and $\boldsymbol{v}_{clip}$ obtained from the image encoder $\textbf{F}_{clip}^{img}$ of CLIP, into the interaction decoder. Tab.~\ref{tab:representations} demonstrates their effectiveness. Initially, we establish a baseline model following the design of GEN-VLKT~\cite{liao2022gen} while excluding the knowledge distillation component. Subsequently, the incorporation of $\boldsymbol{V}_{sd}$ leads to significant improvements, particularly in terms of rare mAP. This suggests that the internal features in stable diffusion play a crucial role in enhancing the representation of rare categories in the HICO-DET dataset. Additionally, the integration of $\boldsymbol{v}_{clip}$ further aids in improving the non-rare AP, as it aligns the output queries with the interaction classifier.

\textbf{Diffusion Time Steps.} We investigate the effectiveness of different diffusion steps in extracting interaction-aware features $\boldsymbol{V}_{sd}$, similarly to~\cite{baranchuk2021label,xu2023open}. Diffusion models control the noise distortion added to the input image by varying the value of $t$. Stable diffusion~\cite{rombach2022high} uses a total of $1000$ time steps. We set $t$ values to $0, 100, 500$ for ablation studies. As demonstrated in Tab.~\ref{tab:time-step}, the best performance is achieved when $t=0$.  It is worth noting that using $\boldsymbol{V}_{sd}$ extracted from input images with higher noise levels would decrease performance and potentially impact the learning of interactions.

\section{Conclusion}
We investigate the issues of the current HOI detection methods and present a novel scheme, namely \texttt{DiffHOI}, which leverages both the generative and representation capacities of a pre-trained text-to-image diffusion
model to benefit the HOI detection task. Particularly, we release a class-balance, large-scale, and high-diversity synthetic HOI dataset called \texttt{SynHOI} to address the long-tail issue in previous datasets and develop an automatic and scalable pipeline to
scale up the generation of diverse and high-precision HOI-annotated data. Extensive experimental results demonstrate that our method significantly outperforms the prior state-of-the-art in regular and zero-shot detection tasks. We hope this work could inspire further research in related fields.

\bibliography{neurips_2023}

\begin{thebibliography}{66}
\providecommand{\natexlab}[1]{#1}
\providecommand{\url}[1]{\texttt{#1}}
\expandafter\ifx\csname urlstyle\endcsname\relax
  \providecommand{\doi}[1]{doi: #1}\else
  \providecommand{\doi}{doi: \begingroup \urlstyle{rm}\Url}\fi

\bibitem[Heilbron et~al.(2015)Heilbron, Escorcia, Ghanem, and
  Niebles]{heilbron2015activitynet}
Fabian~Caba Heilbron, Victor Escorcia, Bernard Ghanem, and Juan~Carlos Niebles.
\newblock Activitynet: A large-scale video benchmark for human activity
  understanding.
\newblock In \emph{2015 IEEE conference on computer vision and pattern
  recognition (CVPR)}, pages 961--970. IEEE, 2015.

\bibitem[Yi et~al.(2022)Yi, Zhou, Habermann, Shimada, Golyanik, Theobalt, and
  Xu]{yi2022physical}
Xinyu Yi, Yuxiao Zhou, Marc Habermann, Soshi Shimada, Vladislav Golyanik,
  Christian Theobalt, and Feng Xu.
\newblock Physical inertial poser (pip): Physics-aware real-time human motion
  tracking from sparse inertial sensors.
\newblock In \emph{Proceedings of the IEEE/CVF Conference on Computer Vision
  and Pattern Recognition}, pages 13167--13178, 2022.

\bibitem[Liu et~al.(2018)Liu, Luo, Lian, and Gao]{liu2018future}
Wen Liu, Weixin Luo, Dongze Lian, and Shenghua Gao.
\newblock Future frame prediction for anomaly detection--a new baseline.
\newblock In \emph{Proceedings of the IEEE conference on computer vision and
  pattern recognition}, pages 6536--6545, 2018.

\bibitem[Hou et~al.(2020)Hou, Peng, Qiao, and Tao]{hou2020visual}
Zhi Hou, Xiaojiang Peng, Yu~Qiao, and Dacheng Tao.
\newblock Visual compositional learning for human-object interaction detection.
\newblock In \emph{Computer Vision--ECCV 2020: 16th European Conference,
  Glasgow, UK, August 23--28, 2020, Proceedings, Part XV 16}, pages 584--600.
  Springer, 2020.

\bibitem[Wan et~al.(2019)Wan, Zhou, Liu, Li, and He]{wan2019pose}
Bo~Wan, Desen Zhou, Yongfei Liu, Rongjie Li, and Xuming He.
\newblock Pose-aware multi-level feature network for human object interaction
  detection.
\newblock In \emph{Proceedings of the IEEE/CVF International Conference on
  Computer Vision}, pages 9469--9478, 2019.

\bibitem[Liao et~al.(2022)Liao, Zhang, Lu, Wang, Li, and Liu]{liao2022gen}
Yue Liao, Aixi Zhang, Miao Lu, Yongliang Wang, Xiaobo Li, and Si~Liu.
\newblock Gen-vlkt: Simplify association and enhance interaction understanding
  for hoi detection.
\newblock In \emph{Proceedings of the IEEE/CVF Conference on Computer Vision
  and Pattern Recognition}, pages 20123--20132, 2022.

\bibitem[Wu et~al.(2022{\natexlab{a}})Wu, Li, Liu, Zhang, Wu, and
  Lu]{wu2022mining}
Xiaoqian Wu, Yong-Lu Li, Xinpeng Liu, Junyi Zhang, Yuzhe Wu, and Cewu Lu.
\newblock Mining cross-person cues for body-part interactiveness learning in
  hoi detection.
\newblock In \emph{Proceedings of the European Conference on Computer Vision},
  pages 121--136. Springer, 2022{\natexlab{a}}.

\bibitem[Zhang et~al.(2021{\natexlab{a}})Zhang, Liao, Liu, Lu, Wang, Gao, and
  Li]{zhang2021mining}
Aixi Zhang, Yue Liao, Si~Liu, Miao Lu, Yongliang Wang, Chen Gao, and Xiaobo Li.
\newblock Mining the benefits of two-stage and one-stage hoi detection.
\newblock \emph{Advances in Neural Information Processing Systems},
  34:\penalty0 17209--17220, 2021{\natexlab{a}}.

\bibitem[Zhou et~al.(2022{\natexlab{a}})Zhou, Liu, Wang, Wang, Hu, Ding, and
  Wang]{zhou2022human}
Desen Zhou, Zhichao Liu, Jian Wang, Leshan Wang, Tao Hu, Errui Ding, and
  Jingdong Wang.
\newblock Human-object interaction detection via disentangled transformer.
\newblock In \emph{Proceedings of the IEEE/CVF Conference on Computer Vision
  and Pattern Recognition}, pages 19568--19577, 2022{\natexlab{a}}.

\bibitem[Chao et~al.(2018{\natexlab{a}})Chao, Liu, Liu, Zeng, and
  Deng]{chao:wacv2018}
Yu-Wei Chao, Yunfan Liu, Xieyang Liu, Huayi Zeng, and Jia Deng.
\newblock Learning to detect human-object interactions.
\newblock In \emph{Proceedings of the IEEE Winter Conference on Applications of
  Computer Vision}, 2018{\natexlab{a}}.

\bibitem[Chao et~al.(2018{\natexlab{b}})Chao, Liu, Liu, Zeng, and
  Deng]{chao2018learning}
Yu-Wei Chao, Yunfan Liu, Xieyang Liu, Huayi Zeng, and Jia Deng.
\newblock Learning to detect human-object interactions.
\newblock In \emph{2018 ieee winter conference on applications of computer
  vision (wacv)}, pages 381--389. IEEE, 2018{\natexlab{b}}.

\bibitem[Gao et~al.(2018)Gao, Zou, and Huang]{gao2018ican}
Chen Gao, Yuliang Zou, and Jia-Bin Huang.
\newblock ican: Instance-centric attention network for human-object interaction
  detection.
\newblock \emph{arXiv preprint arXiv:1808.10437}, 2018.

\bibitem[Gupta et~al.(2019)Gupta, Schwing, and Hoiem]{gupta2019no}
Tanmay Gupta, Alexander Schwing, and Derek Hoiem.
\newblock No-frills human-object interaction detection: Factorization, layout
  encodings, and training techniques.
\newblock In \emph{Proceedings of the IEEE/CVF International Conference on
  Computer Vision}, pages 9677--9685, 2019.

\bibitem[Liao et~al.(2020)Liao, Liu, Wang, Chen, Qian, and Feng]{liao2020ppdm}
Yue Liao, Si~Liu, Fei Wang, Yanjie Chen, Chen Qian, and Jiashi Feng.
\newblock Ppdm: Parallel point detection and matching for real-time
  human-object interaction detection.
\newblock In \emph{Proceedings of the IEEE/CVF Conference on Computer Vision
  and Pattern Recognition}, pages 482--490, 2020.

\bibitem[Wang et~al.(2020{\natexlab{a}})Wang, Yang, Danelljan, Khan, Zhang, and
  Sun]{wang2020learning}
Tiancai Wang, Tong Yang, Martin Danelljan, Fahad~Shahbaz Khan, Xiangyu Zhang,
  and Jian Sun.
\newblock Learning human-object interaction detection using interaction points.
\newblock In \emph{Proceedings of the IEEE/CVF Conference on Computer Vision
  and Pattern Recognition}, pages 4116--4125, 2020{\natexlab{a}}.

\bibitem[Kim et~al.(2020)Kim, Choi, Kang, and Kim]{kim2020uniondet}
Bumsoo Kim, Taeho Choi, Jaewoo Kang, and Hyunwoo~J Kim.
\newblock Uniondet: Union-level detector towards real-time human-object
  interaction detection.
\newblock In \emph{Computer Vision--ECCV 2020: 16th European Conference,
  Glasgow, UK, August 23--28, 2020, Proceedings, Part XV 16}, pages 498--514.
  Springer, 2020.

\bibitem[Balaji et~al.(2022)Balaji, Nah, Huang, Vahdat, Song, Kreis, Aittala,
  Aila, Laine, Catanzaro, et~al.]{balaji2022ediffi}
Yogesh Balaji, Seungjun Nah, Xun Huang, Arash Vahdat, Jiaming Song, Karsten
  Kreis, Miika Aittala, Timo Aila, Samuli Laine, Bryan Catanzaro, et~al.
\newblock ediffi: Text-to-image diffusion models with an ensemble of expert
  denoisers.
\newblock \emph{arXiv preprint arXiv:2211.01324}, 2022.

\bibitem[Zhou et~al.(2022{\natexlab{b}})Zhou, Zhang, Chen, Li, Tensmeyer, Yu,
  Gu, Xu, and Sun]{zhou2022towards}
Yufan Zhou, Ruiyi Zhang, Changyou Chen, Chunyuan Li, Chris Tensmeyer, Tong Yu,
  Jiuxiang Gu, Jinhui Xu, and Tong Sun.
\newblock Towards language-free training for text-to-image generation.
\newblock In \emph{Proceedings of the IEEE/CVF Conference on Computer Vision
  and Pattern Recognition}, pages 17907--17917, 2022{\natexlab{b}}.

\bibitem[Saharia et~al.(2022)Saharia, Chan, Saxena, Li, Whang, Denton,
  Ghasemipour, Gontijo~Lopes, Karagol~Ayan, Salimans,
  et~al.]{saharia2022photorealistic}
Chitwan Saharia, William Chan, Saurabh Saxena, Lala Li, Jay Whang, Emily~L
  Denton, Kamyar Ghasemipour, Raphael Gontijo~Lopes, Burcu Karagol~Ayan, Tim
  Salimans, et~al.
\newblock Photorealistic text-to-image diffusion models with deep language
  understanding.
\newblock \emph{Advances in Neural Information Processing Systems},
  35:\penalty0 36479--36494, 2022.

\bibitem[Ramesh et~al.(2022)Ramesh, Dhariwal, Nichol, Chu, and
  Chen]{ramesh2022hierarchical}
Aditya Ramesh, Prafulla Dhariwal, Alex Nichol, Casey Chu, and Mark Chen.
\newblock Hierarchical text-conditional image generation with clip latents.
\newblock \emph{arXiv preprint arXiv:2204.06125}, 2022.

\bibitem[Rombach et~al.(2022)Rombach, Blattmann, Lorenz, Esser, and
  Ommer]{rombach2022high}
Robin Rombach, Andreas Blattmann, Dominik Lorenz, Patrick Esser, and Bj{\"o}rn
  Ommer.
\newblock High-resolution image synthesis with latent diffusion models.
\newblock In \emph{Proceedings of the IEEE/CVF Conference on Computer Vision
  and Pattern Recognition}, pages 10684--10695, 2022.

\bibitem[Tang et~al.(2022)Tang, Pandey, Jiang, Yang, Kumar, Lin, and
  Ture]{tang2022daam}
Raphael Tang, Akshat Pandey, Zhiying Jiang, Gefei Yang, Karun Kumar, Jimmy Lin,
  and Ferhan Ture.
\newblock What the daam: Interpreting stable diffusion using cross attention.
\newblock \emph{arXiv preprint arXiv:2210.04885}, 2022.

\bibitem[Xu et~al.(2023)Xu, Liu, Vahdat, Byeon, Wang, and De~Mello]{xu2023open}
Jiarui Xu, Sifei Liu, Arash Vahdat, Wonmin Byeon, Xiaolong Wang, and Shalini
  De~Mello.
\newblock Open-vocabulary panoptic segmentation with text-to-image diffusion
  models.
\newblock \emph{arXiv preprint arXiv:2303.04803}, 2023.

\bibitem[Li et~al.(2023)Li, Zhou, Zhang, Zhang, Wang, and Xie]{li2023guiding}
Ziyi Li, Qinye Zhou, Xiaoyun Zhang, Ya~Zhang, Yanfeng Wang, and Weidi Xie.
\newblock Guiding text-to-image diffusion model towards grounded generation.
\newblock \emph{arXiv preprint arXiv:2301.05221}, 2023.

\bibitem[Shipard et~al.()Shipard, Wiliem, Thanh, Xiang, and
  Fookes]{shiparddiversity}
Jordan Shipard, Arnold Wiliem, Kien~Nguyen Thanh, Wei Xiang, and Clinton
  Fookes.
\newblock Diversity is definitely needed: Improving model-agnostic zero-shot
  classification via stable diffusion.

\bibitem[Li et~al.(2019)Li, Zhou, Huang, Xu, Ma, Fang, Wang, and
  Lu]{li2019transferable}
Yong-Lu Li, Siyuan Zhou, Xijie Huang, Liang Xu, Ze~Ma, Hao-Shu Fang, Yanfeng
  Wang, and Cewu Lu.
\newblock Transferable interactiveness knowledge for human-object interaction
  detection.
\newblock In \emph{Proceedings of the IEEE/CVF Conference on Computer Vision
  and Pattern Recognition}, pages 3585--3594, 2019.

\bibitem[Hou et~al.(2021{\natexlab{a}})Hou, Yu, Qiao, Peng, and
  Tao]{hou2021detecting}
Zhi Hou, Baosheng Yu, Yu~Qiao, Xiaojiang Peng, and Dacheng Tao.
\newblock Detecting human-object interaction via fabricated compositional
  learning.
\newblock In \emph{Proceedings of the IEEE/CVF Conference on Computer Vision
  and Pattern Recognition}, pages 14646--14655, 2021{\natexlab{a}}.

\bibitem[Xu et~al.(2019)Xu, Wong, Li, Zhao, and Kankanhalli]{xu2019learning}
Bingjie Xu, Yongkang Wong, Junnan Li, Qi~Zhao, and Mohan~S Kankanhalli.
\newblock Learning to detect human-object interactions with knowledge.
\newblock In \emph{Proceedings of the IEEE/CVF Conference on Computer Vision
  and Pattern Recognition}, 2019.

\bibitem[Bansal et~al.(2020)Bansal, Rambhatla, Shrivastava, and
  Chellappa]{bansal2020detecting}
Ankan Bansal, Sai~Saketh Rambhatla, Abhinav Shrivastava, and Rama Chellappa.
\newblock Detecting human-object interactions via functional generalization.
\newblock In \emph{Proceedings of the AAAI Conference on Artificial
  Intelligence}, volume~34, pages 10460--10469, 2020.

\bibitem[Li et~al.(2020)Li, Liu, Lu, Wang, Liu, Li, and Lu]{li2020detailed}
Yong-Lu Li, Xinpeng Liu, Han Lu, Shiyi Wang, Junqi Liu, Jiefeng Li, and Cewu
  Lu.
\newblock Detailed 2d-3d joint representation for human-object interaction.
\newblock In \emph{Proceedings of the IEEE/CVF Conference on Computer Vision
  and Pattern Recognition}, pages 10166--10175, 2020.

\bibitem[Liu et~al.(2020)Liu, Yuan, and Chen]{liu2020consnet}
Ye~Liu, Junsong Yuan, and Chang~Wen Chen.
\newblock Consnet: Learning consistency graph for zero-shot human-object
  interaction detection.
\newblock In \emph{Proceedings of the 28th ACM International Conference on
  Multimedia}, pages 4235--4243, 2020.

\bibitem[Gao et~al.(2020)Gao, Xu, Zou, and Huang]{gao2020drg}
Chen Gao, Jiarui Xu, Yuliang Zou, and Jia-Bin Huang.
\newblock Drg: Dual relation graph for human-object interaction detection.
\newblock In \emph{Computer Vision--ECCV 2020: 16th European Conference,
  Glasgow, UK, August 23--28, 2020, Proceedings, Part XII 16}, pages 696--712.
  Springer, 2020.

\bibitem[Qi et~al.(2018)Qi, Wang, Jia, Shen, and Zhu]{qi2018learning}
Siyuan Qi, Wenguan Wang, Baoxiong Jia, Jianbing Shen, and Song-Chun Zhu.
\newblock Learning human-object interactions by graph parsing neural networks.
\newblock In \emph{Proceedings of the European conference on computer vision
  (ECCV)}, pages 401--417, 2018.

\bibitem[Wang et~al.(2020{\natexlab{b}})Wang, Zheng, and
  Yingbiao]{wang2020contextual}
Hai Wang, Wei-shi Zheng, and Ling Yingbiao.
\newblock Contextual heterogeneous graph network for human-object interaction
  detection.
\newblock In \emph{Computer Vision--ECCV 2020: 16th European Conference,
  Glasgow, UK, August 23--28, 2020, Proceedings, Part XVII 16}, pages 248--264.
  Springer, 2020{\natexlab{b}}.

\bibitem[Zhang et~al.(2021{\natexlab{b}})Zhang, Campbell, and
  Gould]{zhang2021spatially}
Frederic~Z Zhang, Dylan Campbell, and Stephen Gould.
\newblock Spatially conditioned graphs for detecting human-object interactions.
\newblock In \emph{Proceedings of the IEEE/CVF International Conference on
  Computer Vision}, pages 13319--13327, 2021{\natexlab{b}}.

\bibitem[Park et~al.(2023)Park, Park, and Lee]{park2023viplo}
Jeeseung Park, Jin-Woo Park, and Jong-Seok Lee.
\newblock Viplo: Vision transformer based pose-conditioned self-loop graph for
  human-object interaction detection.
\newblock \emph{arXiv preprint arXiv:2304.08114}, 2023.

\bibitem[Fang et~al.(2021)Fang, Xie, Shao, and Lu]{fang2021dirv}
Hao-Shu Fang, Yichen Xie, Dian Shao, and Cewu Lu.
\newblock Dirv: Dense interaction region voting for end-to-end human-object
  interaction detection.
\newblock In \emph{Proceedings of the AAAI Conference on Artificial
  Intelligence}, volume~35, pages 1291--1299, 2021.

\bibitem[Tamura et~al.(2021{\natexlab{a}})Tamura, Ohashi, and
  Yoshinaga]{Tamura_2021_CVPR}
Masato Tamura, Hiroki Ohashi, and Tomoaki Yoshinaga.
\newblock Qpic: Query-based pairwise human-object interaction detection with
  image-wide contextual information.
\newblock In \emph{Proceedings of the IEEE/CVF Conference on Computer Vision
  and Pattern Recognition (CVPR)}, pages 10410--10419, June 2021{\natexlab{a}}.

\bibitem[Tamura et~al.(2021{\natexlab{b}})Tamura, Ohashi, and
  Yoshinaga]{tamura2021qpic}
Masato Tamura, Hiroki Ohashi, and Tomoaki Yoshinaga.
\newblock Qpic: Query-based pairwise human-object interaction detection with
  image-wide contextual information.
\newblock In \emph{Proceedings of the IEEE/CVF Conference on Computer Vision
  and Pattern Recognition}, pages 10410--10419, 2021{\natexlab{b}}.

\bibitem[Kim et~al.(2021)Kim, Lee, Kang, Kim, and Kim]{kim2021hotr}
Bumsoo Kim, Junhyun Lee, Jaewoo Kang, Eun-Sol Kim, and Hyunwoo~J Kim.
\newblock Hotr: End-to-end human-object interaction detection with
  transformers.
\newblock In \emph{Proceedings of the IEEE/CVF Conference on Computer Vision
  and Pattern Recognition}, pages 74--83, 2021.

\bibitem[Chen and Yanai(2021)]{chen2021qahoi}
Junwen Chen and Keiji Yanai.
\newblock Qahoi: query-based anchors for human-object interaction detection.
\newblock \emph{arXiv preprint arXiv:2112.08647}, 2021.

\bibitem[Ma et~al.(2023)Ma, Wang, Wang, and Wei]{ma2023fgahoi}
Shuailei Ma, Yuefeng Wang, Shanze Wang, and Ying Wei.
\newblock Fgahoi: Fine-grained anchors for human-object interaction detection.
\newblock \emph{arXiv preprint arXiv:2301.04019}, 2023.

\bibitem[Chen et~al.(2021)Chen, Liao, Liu, Chen, Wang, and
  Qian]{chen2021reformulating}
Mingfei Chen, Yue Liao, Si~Liu, Zhiyuan Chen, Fei Wang, and Chen Qian.
\newblock Reformulating hoi detection as adaptive set prediction.
\newblock In \emph{Proceedings of the IEEE/CVF Conference on Computer Vision
  and Pattern Recognition}, pages 9004--9013, 2021.

\bibitem[Peyre et~al.(2019)Peyre, Laptev, Schmid, and
  Sivic]{peyre2019detecting}
Julia Peyre, Ivan Laptev, Cordelia Schmid, and Josef Sivic.
\newblock Detecting unseen visual relations using analogies.
\newblock In \emph{Proceedings of the IEEE/CVF International Conference on
  Computer Vision}, pages 1981--1990, 2019.

\bibitem[Hou et~al.(2021{\natexlab{b}})Hou, Yu, Qiao, Peng, and
  Tao]{hou2021affordance}
Zhi Hou, Baosheng Yu, Yu~Qiao, Xiaojiang Peng, and Dacheng Tao.
\newblock Affordance transfer learning for human-object interaction detection.
\newblock In \emph{Proceedings of the IEEE/CVF Conference on Computer Vision
  and Pattern Recognition}, pages 495--504, 2021{\natexlab{b}}.

\bibitem[Radford et~al.(2021)Radford, Kim, Hallacy, Ramesh, Goh, Agarwal,
  Sastry, Askell, Mishkin, Clark, et~al.]{radford2021learning}
Alec Radford, Jong~Wook Kim, Chris Hallacy, Aditya Ramesh, Gabriel Goh,
  Sandhini Agarwal, Girish Sastry, Amanda Askell, Pamela Mishkin, Jack Clark,
  et~al.
\newblock Learning transferable visual models from natural language
  supervision.
\newblock In \emph{International conference on machine learning}, pages
  8748--8763. PMLR, 2021.

\bibitem[Wu et~al.(2022{\natexlab{b}})Wu, Gu, Shen, Lin, Chen, Sun, and
  Ji]{wu2022end}
Mingrui Wu, Jiaxin Gu, Yunhang Shen, Mingbao Lin, Chao Chen, Xiaoshuai Sun, and
  Rongrong Ji.
\newblock End-to-end zero-shot hoi detection via vision and language knowledge
  distillation.
\newblock \emph{arXiv preprint arXiv:2204.03541}, 2022{\natexlab{b}}.

\bibitem[Ning et~al.(2023)Ning, Qiu, Liu, and He]{ning2023hoiclip}
Shan Ning, Longtian Qiu, Yongfei Liu, and Xuming He.
\newblock Hoiclip: Efficient knowledge transfer for hoi detection with
  vision-language models.
\newblock \emph{arXiv preprint arXiv:2303.15786}, 2023.

\bibitem[Trabucco et~al.(2023)Trabucco, Doherty, Gurinas, and
  Salakhutdinov]{trabucco2023effective}
Brandon Trabucco, Kyle Doherty, Max Gurinas, and Ruslan Salakhutdinov.
\newblock Effective data augmentation with diffusion models.
\newblock \emph{arXiv preprint arXiv:2302.07944}, 2023.

\bibitem[Bansal and Grover(2023)]{bansal2023leaving}
Hritik Bansal and Aditya Grover.
\newblock Leaving reality to imagination: Robust classification via generated
  datasets.
\newblock \emph{arXiv preprint arXiv:2302.02503}, 2023.

\bibitem[You et~al.(2023)You, Zhong, Bao, Sun, Li, and Zhu]{you2023diffusion}
Zebin You, Yong Zhong, Fan Bao, Jiacheng Sun, Chongxuan Li, and Jun Zhu.
\newblock Diffusion models and semi-supervised learners benefit mutually with
  few labels.
\newblock \emph{arXiv preprint arXiv:2302.10586}, 2023.

\bibitem[Zhang et~al.(2022{\natexlab{a}})Zhang, Li, Liu, Zhang, Su, Zhu, Ni,
  and Shum]{zhang2022dino}
Hao Zhang, Feng Li, Shilong Liu, Lei Zhang, Hang Su, Jun Zhu, Lionel~M Ni, and
  Heung-Yeung Shum.
\newblock Dino: Detr with improved denoising anchor boxes for end-to-end object
  detection.
\newblock \emph{arXiv preprint arXiv:2203.03605}, 2022{\natexlab{a}}.

\bibitem[Lin et~al.(2014)Lin, Maire, Belongie, Hays, Perona, Ramanan,
  Doll{\'a}r, and Zitnick]{lin2014microsoft}
Tsung-Yi Lin, Michael Maire, Serge Belongie, James Hays, Pietro Perona, Deva
  Ramanan, Piotr Doll{\'a}r, and C~Lawrence Zitnick.
\newblock Microsoft coco: Common objects in context.
\newblock In \emph{Computer Vision--ECCV 2014: 13th European Conference,
  Zurich, Switzerland, September 6-12, 2014, Proceedings, Part V 13}, pages
  740--755. Springer, 2014.

\bibitem[Hessel et~al.(2021)Hessel, Holtzman, Forbes, Bras, and
  Choi]{hessel2021clipscore}
Jack Hessel, Ari Holtzman, Maxwell Forbes, Ronan~Le Bras, and Yejin Choi.
\newblock Clipscore: A reference-free evaluation metric for image captioning.
\newblock \emph{arXiv preprint arXiv:2104.08718}, 2021.

\bibitem[Carion et~al.(2020)Carion, Massa, Synnaeve, Usunier, Kirillov, and
  Zagoruyko]{carion2020end}
Nicolas Carion, Francisco Massa, Gabriel Synnaeve, Nicolas Usunier, Alexander
  Kirillov, and Sergey Zagoruyko.
\newblock End-to-end object detection with transformers.
\newblock In \emph{Proceedings of the European Conference on Computer Vision},
  pages 213--229. Springer, 2020.

\bibitem[Gupta and Malik(2015)]{gupta2015visual}
Saurabh Gupta and Jitendra Malik.
\newblock Visual semantic role labeling.
\newblock \emph{arXiv preprint arXiv:1505.04474}, 2015.

\bibitem[Liu et~al.(2021)Liu, Lin, Cao, Hu, Wei, Zhang, Lin, and
  Guo]{liu2021swin}
Ze~Liu, Yutong Lin, Yue Cao, Han Hu, Yixuan Wei, Zheng Zhang, Stephen Lin, and
  Baining Guo.
\newblock Swin transformer: Hierarchical vision transformer using shifted
  windows.
\newblock In \emph{Proceedings of the IEEE/CVF international conference on
  computer vision}, pages 10012--10022, 2021.

\bibitem[Zhu et~al.(2020)Zhu, Su, Lu, Li, Wang, and Dai]{zhu2020deformable}
Xizhou Zhu, Weijie Su, Lewei Lu, Bin Li, Xiaogang Wang, and Jifeng Dai.
\newblock Deformable detr: Deformable transformers for end-to-end object
  detection.
\newblock \emph{arXiv preprint arXiv:2010.04159}, 2020.

\bibitem[Zhang et~al.(2022{\natexlab{b}})Zhang, Pan, Yao, Huang, Mei, and
  Chen]{zhang2022exploring}
Yong Zhang, Yingwei Pan, Ting Yao, Rui Huang, Tao Mei, and Chang-Wen Chen.
\newblock Exploring structure-aware transformer over interaction proposals for
  human-object interaction detection.
\newblock In \emph{Proceedings of the IEEE/CVF Conference on Computer Vision
  and Pattern Recognition}, pages 19548--19557, 2022{\natexlab{b}}.

\bibitem[Jin et~al.(2022)Jin, Chen, Wang, Wang, Yu, Liang, Hwang, and
  Liu]{jin2022overlooked}
Ying Jin, Yinpeng Chen, Lijuan Wang, Jianfeng Wang, Pei Yu, Lin Liang,
  Jenq-Neng Hwang, and Zicheng Liu.
\newblock The overlooked classifier in human-object interaction recognition.
\newblock \emph{arXiv preprint arXiv:2203.05676}, 2022.

\bibitem[Zhang et~al.(2022{\natexlab{c}})Zhang, Campbell, and
  Gould]{zhang2022efficient}
Frederic~Z Zhang, Dylan Campbell, and Stephen Gould.
\newblock Efficient two-stage detection of human-object interactions with a
  novel unary-pairwise transformer.
\newblock In \emph{Proceedings of the IEEE/CVF Conference on Computer Vision
  and Pattern Recognition}, pages 20104--20112, 2022{\natexlab{c}}.

\bibitem[Kim et~al.(2022)Kim, Mun, On, Shin, Lee, and Kim]{kim2022mstr}
Bumsoo Kim, Jonghwan Mun, Kyoung-Woon On, Minchul Shin, Junhyun Lee, and
  Eun-Sol Kim.
\newblock Mstr: Multi-scale transformer for end-to-end human-object interaction
  detection.
\newblock In \emph{Proceedings of the IEEE/CVF Conference on Computer Vision
  and Pattern Recognition}, pages 19578--19587, 2022.

\bibitem[Iftekhar et~al.(2022)Iftekhar, Chen, Kundu, Li, Tighe, and
  Modolo]{iftekhar2022look}
ASM Iftekhar, Hao Chen, Kaustav Kundu, Xinyu Li, Joseph Tighe, and Davide
  Modolo.
\newblock What to look at and where: Semantic and spatial refined transformer
  for detecting human-object interactions.
\newblock In \emph{Proceedings of the IEEE/CVF Conference on Computer Vision
  and Pattern Recognition}, pages 5353--5363, 2022.

\bibitem[Qu et~al.(2022)Qu, Ding, Li, Zhong, and Tao]{qu2022distillation}
Xian Qu, Changxing Ding, Xingao Li, Xubin Zhong, and Dacheng Tao.
\newblock Distillation using oracle queries for transformer-based human-object
  interaction detection.
\newblock In \emph{Proceedings of the IEEE/CVF Conference on Computer Vision
  and Pattern Recognition}, pages 19558--19567, 2022.

\bibitem[Liu et~al.(2022)Liu, Li, Wu, Tai, Lu, and
  Tang]{liu2022interactiveness}
Xinpeng Liu, Yong-Lu Li, Xiaoqian Wu, Yu-Wing Tai, Cewu Lu, and Chi-Keung Tang.
\newblock Interactiveness field in human-object interactions.
\newblock In \emph{Proceedings of the IEEE/CVF Conference on Computer Vision
  and Pattern Recognition}, pages 20113--20122, 2022.

\bibitem[Baranchuk et~al.(2021)Baranchuk, Rubachev, Voynov, Khrulkov, and
  Babenko]{baranchuk2021label}
Dmitry Baranchuk, Ivan Rubachev, Andrey Voynov, Valentin Khrulkov, and Artem
  Babenko.
\newblock Label-efficient semantic segmentation with diffusion models.
\newblock \emph{arXiv preprint arXiv:2112.03126}, 2021.

\end{thebibliography}
\bibliographystyle{unsrtnat}

\end{document}